# Annealed MAP

**Changhe Yuan**
Decision Systems Laboratory
Intelligent Systems Program
University of Pittsburgh
Pittsburgh, PA 15260
`cyuan@sis.pitt.edu`

**Tsai-Ching Lu**
HRL Laboratories, LLC
Malibu, CA 90265
`tlu@hrl.com`

**Marek J. Druzdzel**
Decision Systems Laboratory
School of Information Sciences and
Intelligent Systems Program
University of Pittsburgh
Pittsburgh, PA 15260
`marek@sis.pitt.edu`

## Abstract

Maximum a Posteriori assignment (MAP) is the problem of finding the most probable instantiation of a set of variables given the partial evidence on the other variables in a Bayesian network. MAP has been shown to be a NP-hard problem [22], even for constrained networks, such as polytrees [18]. Hence, previous approaches often fail to yield any results for MAP problems in large complex Bayesian networks. To address this problem, we propose ANNEALEDMAP algorithm, a simulated annealing-based MAP algorithm. The ANNEALEDMAP algorithm simulates a non-homogeneous Markov chain whose invariant function is a probability density that concentrates itself on the modes of the target density. We tested this algorithm on several real Bayesian networks. The results show that, while maintaining good quality of the MAP solutions, the ANNEALEDMAP algorithm is also able to solve many problems that are beyond the reach of previous approaches.

## 1  Introduction

MAP is the problem of finding the most probable instantiation of one set of variables given partial evidence on the remaining variables in a Bayesian network. One special case of MAP is the Most Probable Explanation (MPE) problem. MPE is the problem of finding the most probable instantiation of a set of variables given full evidence on the remaining variables. Due to its simplicity, MPE has received much more attention than MAP. However, practice tells us that MAP has much greater value in real problems. For instance, we often want to know the most probable instantiation of some target nodes[1] given the states of some of the observable nodes in Bayesian networks. Researchers have proposed various approaches to solve the MAP problem [6, 7, 8, 19, 20]. The state of the art MAP algorithm, proposed in a recent paper by Park and Darwiche [20], is a branch-and-bound depth-first search algorithm. It is an efficient algorithm, especially when the search spaces of the problems are not too large. However, when we applied the algorithm to MAP problems in some large real Bayesian networks, it failed to yield any result. Given that MAP has been shown to be an NP-hard problem [22], this is not surprising. Furthermore, MAP is NP-hard even for constrained networks, such as polytrees [18].

In this paper, we propose the ANNEALEDMAP algorithm, which uses Markov Chain Monte Carlo (MCMC) methods to sample from the target distribution $p(x)$, and applies the simulated annealing technique to simulate a non-homogeneous Markov chain whose invariant function is $p^{1/T_i}(X|E)$, which is a probability density that concentrates itself on the modes of $p(x)$ as $\lim_{i \to \infty} T_i = 0$.

The remaining of this paper is organized as follows. In Section 2, we define the MAP problem more formally and analyze why it is difficult. We also outline several existing approaches. In Section 3, we propose the ANNEALEDMAP algorithm. First, we introduce the MCMC methods. Second, we introduce the simulated annealing (SA) technique, which is often used to solve global optimization problem. After that, we combine these two techniques into the ANNEALEDMAP algorithm to solve the MAP problem in Bayesian networks. In Section 4, we present the results of applying the ANNEALEDMAP algorithm to several real complex Bayesian networks.

---

[1]In this paper, variable and node are used interchangeably, which is also the case for configuration and instantiation.

## 2 Previous Work

Formally, the MAP problem in a Bayesian network is defined as follows. Let $X$ be the set of MAP nodes whose most probable configuration we are interested in. Let $E$ be the set of evidence nodes. The remainder of the nodes are denoted by $Y$. Let $Z = X \cup Y \cup E$. Then, MAP is the instantiation that satisfies

$$\widehat{x} = max_X p(X|E) . \qquad (1)$$

Using the law of total probability and the chain rule for Bayesian networks, we have

$$\widehat{x} = max_X \sum_Y p(X, Y|E) \qquad (2)$$

$$= max_X \sum_Y \prod_{Z_i \in Z} p(Z_i|PA(Z_i)) , \qquad (3)$$

where $PA(Z_i)$ stands for the parents of node $Z_i$. From Equation 3, we can easily see the difference between MAP and other inference problems. In computing posterior marginal distributions, we only have summations. Thus, we can commute summations over different variables in order to minimize the width of an elimination order. The width of an elimination order is defined as the size of the largest clique minus 1 in a junction tree constructed based on the elimination order.[2] Similarly, we have only maximizations in a MPE problem. Once again, any permutation of the maximizations over different variables is admissible. However, a MAP problem has both maximizations and summations. Since summation and maximization do not commute, we are required to do summations first. An elimination order is valid [18] if maximizing a variable out of a potential never happens before summing over another variable on the same potential. A MAP problem is subject to the constrained width of the best valid elimination order. It is still possible to find valid orderings that interleave summation and maximization variables. However, Park shows that there is always an elimination order with the same width in which all the maximizations are done last, hence, there is no benefit of interleaving summations and maximizations [18].

To solve the MAP problem for Bayesian networks, researchers have proposed various approaches, all of which are trying to sidestep its inherent complexity. The approach in [6] uses genetic algorithms to approximate the best configuration of the MAP variables.

Starting from an initial guess, the algorithm takes actions like crossover and mutation to explore the space of possible instantiations. It stops when a fixed number of iterations have been executed and then choose the best instantiation as the MAP solution. Dechter and Rish [8] propose a general scheme for probabilistic inference: Mini-buckets. A full mini-bucket algorithm is subject to the size of the largest potential created, which is equivalent to the constrained width of the MAP problem. Hence, the mini-bucket method sets a limit on the size of potentials. Whenever the size of a potential exceeds the limit, the mini-bucket method will create an approximate version of it instead. Park and Darwiche [19] propose an approach using local search to solve a MAP problem. The algorithm starts from an initial guess and then iteratively improves the solution by moving to a better neighbor. In a later paper [20], the authors improve the local search algorithm using a branch-and-bound depth-first algorithm. The advantage of the improved algorithm is that it provides a guarantee on the optimality of the obtained solution.

All the above approaches alleviate to some degree the complexity of the original problems. However, in face of large complex models, they often fail to provide good results, if any: the approach in [6] does not provide any guidance to explore the more probable spaces; the quality of the results of the mini-bucket method largely depends on the limit of the potential size; the algorithms in [19, 20] reduce the complexity of the MAP problems to treewidths, but they are still subject to the exponential search spaces introduced in the problems. In this paper, we propose a Monte Carlo sampling based MAP algorithm, which not only reduces the complexity of the problems by sampling, but also cuts the search space by simulated annealing. Similar ideas have been applied to improving the results in [6] and to image restoration in Markov Random Fields (MRF) [7, 10].

## 3 The ANNEALEDMAP Algorithm

One effective way of dealing with the complexity of an inference problem is to use Monte Carlo sampling algorithms. Sampling-based algorithms are inevitably slower than exact algorithms when the problems at hand are easy; however, they largely reduce the complexity of the problems by trading off some precision. When the problems become too complex, they may be the only feasible approaches. Furthermore, they have the nice property of being anytime algorithms. In this section, we propose the ANNEALEDMAP algorithm, which integrates the Markov Chain Monte Carlo methods and the simulated annealing technique to solve MAP problems in Bayesian networks.

---

[2]The width of the best elimination order is called the network treewidth. The constrained width is defined as the width of the best constrained elimination order with respect to a specific MAP problem.

This section is organized as follows. First, we introduce the Markov Chain Monte Carlo methods. Second, we introduce the simulated annealing technique, which is often used to solve global optimization problems. After that, we discuss how to combine them into the ANNEALEDMAP algorithm, which aims to solve the MAP problems in Bayesian networks. Finally, since annealing schedules are so important for the simulated annealing technique, we discuss the annealing schedule that we are using.

### 3.1 Markov Chain Monte Carlo Methods

MCMC methods [4] are techniques to sample from the target density $p(x)$ by constructing a Markov chain with $p(x)$ as the invariant function. Let $T$ be the transition matrix of the constructed Markov chain. It has been shown that for any starting point, the chain will converge to the invariant function $p(x)$ if $T$ satisfies two properties:

1. Irreducibility: There is a positive transition probability between any two states within a limited number of steps.

2. Aperiodicity: The chain will not be trapped in cycles.

One way to design a Markov chain with transition matrix $T$ that has the target density $p(x)$ as the invariant function is to make sure that $p(x^*)T(x|x^*) = p(x)T(x^*|x)$, which is known as the *detailed balance condition*. However, this condition is only sufficient but not necessary.

The Metropolis-Hasting (MH) algorithm is the most popular MCMC method. The Markov Chain of the MH algorithm moves to a new state $x^*$ with acceptance probability $A(x, x^*) = min\{1, \frac{p(x^*)q(x|x^*)}{p(x)q(x^*|x)}\}$, where $q(x^*|x)$ is the proposal distribution. It is easy to show that the transition kernel defined by $q(x^*|x)$ satisfies the *detailed balance condition*. Gibbs sampling is one special case of MH algorithm. Suppose we have n-dimensional vector $X = x_1, x_2, ..., x_n$ and the expressions for the full conditionals $p(x_j|x_{-j})$ for each $j$, where $X_{-j} = x_1, ..., x_{j-1}, x_{j+1}, ..., x_n$. Gibbs sampling uses the following proposal distribution for $j = 1, ..., n$

$$q(x^*|x) = \begin{cases} p(x_j^*|x_{-j}), & \text{if } x_{-j}^* = x_{-j} \text{ ;} \\ 0, & \text{otherwise .} \end{cases}$$

It is easy to verify that acceptance probability is always 1 for Gibbs sampling.

### 3.2 Simulated Annealing for Global Optimization

Now we already have the techniques to sample from the target density $p(x)$. Moreover, if we want to find its global maximum, we can use simulated annealing technique [13] to simulate a non-homogeneous Markov chain whose invariant function is no longer $p(x)$, but

$$p_i(x) \propto p^{1/T_i}(x) \, , \tag{4}$$

Where $\lim_{i \to \infty} T_i = 0$. Under weak regularity assumption, $p^{\infty}(x)$ is a probability density that concentrates itself on the modes of $p(x)$.

### 3.3 The ANNEALEDMAP Algorithm

Now let us look at the MAP problem in a Bayesian network. We want to find the most probable configuration of one set of variables $X$ given another set of evidence variables $E$ in the Bayesian network, i.e., $max_X p(X|E)$. Given the technique we described above, it is sufficient to simulate a Markov chain whose invariant function is $p^{1/T_i}(X|E)$. We propose to use the following proposal distribution

$$q_i(x^*|x) = \begin{cases} p^{1/T_i}(x_j^*|x_{-j}), & \text{if } x_{-j}^* = x_{-j} \text{ ;} \\ 0, & \text{otherwise .} \end{cases}$$

The corresponding acceptance probability is

$$\begin{aligned} A(x^*|x) &= min\{1, \frac{p^{1/T_i}(x^*|E)q(x|x^*)}{p^{1/T_i}(x|E)q(x^*|x)}\} \\ &= min\{1, \frac{p^{1/T_i}(x_j^*, x_{-j}^*|E)p(x_j|x_{-j}^*, E)}{p^{1/T_i}(x_j, x_{-j}|E)p(x_j^*|x_{-j}, E)}\} \\ &= min\{1, \frac{p^{1/T_i}(x_j^*|x_{-j}^*, E)p^{1/T_i}(x_{-j}^*|E)p(x_j|x_{-j}^*, E)}{p^{1/T_i}(x_j|x_{-j}, E)p^{1/T_i}(x_{-j}|E)p(x_j^*|x_{-j}, E)}\} \\ &= min\{1, \frac{p^{1/T_i - 1}(x_j^*|x_{-j}, E)}{p^{1/T_i - 1}(x_j|x_{-j}, E)}\} \, . \end{aligned}$$

Thus, calculating the acceptance probability reduces to the local calculation of the posterior distribution of a single node. When $T_i$ is 1, the above algorithm is just Gibbs sampling with invariant function $p(X|E)$. By cooling down $T_i$ to 0, we eventually get a density function which concentrates itself on the global maxima of $p(X|E)$.

To calculate $p(x_j|x_{-j}, E)$, we need to do inference in a Bayesian network. Since exact inference is feasible in the Bayesian networks in our experiments, we use the exact clustering algorithm [15] powered by relevance reasoning [16]. If inference is hard, we can resort to approximate solutions. In such situations, we recommend to use the *Loopy Belief Propagation* algorithm (LBP) [17].

Another important issue is how to initialize the Markov chain. Although theoretically the starting point is not important for sampling from a target distribution with a Markov chain, a good starting point helps a lot when searching for the global maximum of the distribution. In our experiments, we choose the *sequential initialization method*, which chooses each time a MAP node and instantiate it to its most probable state conditioning on the evidence and MAP nodes already initialized.

---

**Algorithm**: ANNEALEDMAP

**Input**: Bayesian network B, a set of MAP variables X, and a set of evidence variables E;

**Output**: The most probable configuration of X.

1. Initialize $\mathbf{X}^{(0)}$, $T_0$, and set $i = 0$.

2. **while** stopping rule is not satisfied:

3.      **for** each variable $x_j$ in the MAP set $X$ do:

4.          Sample $u \sim U_{[0,1]}$.

5.          Sample $x_j^* \sim p(x_j^* | x_{-j}, E)$

6.          if $u < min\{1, \frac{p^{1/T_i-1}(x_j^* | x_{-j}, E)}{p^{1/T_i-1}(x_j | x_{-j}, E)}\}$

             $x_j^{i+1} = x_j^*$.

         else

             $x_j^{i+1} = x_j^i$.

7.          Keep track of the best configuration so far.

     **end for**

8.      Set $T_{i+1}$ according to a chosen annealing schedule.

9.      $i \leftarrow i + 1$.

     **end while**

10. Output the best configuration found and its probability.

---

Figure 1: The ANNEALEDMAP algorithm.

Given the above discussion, we outline the ANNEALEDMAP algorithm in Figure 1. When the annealing schedule is good enough, we can usually take the final configuration as the output. However, since the probability of a configuration can be calculated as a byproduct of the acceptance probability, we can keep track of the best configuration that we have visited. To see this, we have

$$p(x^*|E) \quad = \quad p(x_j^*, x_{-j}^*|E)$$

$$= \quad p(x_{-j}|E)p(x_j^*|x_{-j}, E)$$

$$= \quad p(x|E)\frac{p(x_j^*|x_{-j}, E)}{p(x_j|x_{-j}, E)} \; .$$

### 3.4 Annealing Schedules

While simulated annealing can theoretically converge to the global optimum with stationary distributions, this requires exponential-time execution of the annealing algorithm [13]. In practice, only a short annealing schedule can be used. The annealing schedule that we are using is the *geometric cooling*, where the new temperature $(T')$ is computed using

$$T' = \alpha T \; , \tag{5}$$

where $\alpha$ $(0 < \alpha < 1)$ is the cooling rate, which usually takes a value between 0.8 and 0.99. We stop the annealing algorithm when there has been no improvement for a certain number of iterations. This annealing schedule has the advantage of being well understood, having a solid theoretical foundation, and being the most widely used annealing schedule. However, the schedule also has the drawback that once it drops into a local maximum, it is unlikely to be able to escape it. To overcome this drawback, we empower the *geometric cooling* schedule with another technique, which is called *reheating as a function of cost* (RFC) [9]. This technique involves calculating the specific heat $(C_H)$, which is a measure of the variance of the cost of the values of states at a given temperature. Generally, before the annealing algorithm reaches the temperature $T(C_H)$ when the specific heat is maximal, it is still creating the super-structure of the solution by taking global random walks. After the algorithm reaches $T(C_H)$, it begins to solve the sub-problems. By reheating the system to a temperature above $T(C_H)$, we can escape the local maximum and possibly find better solutions. The specific heat is defined as

$$C_H(T) = \frac{\sigma^2(T)}{T^2} \; , \tag{6}$$

where $\sigma^2(T)$ is the variance of the cost, and $T$ is the current temperature. The score in our experiments is defined as the discrepancy between the current best solution and the current solution. After there has been no improvement for a certain number of iterations, we reheat the system to the new temperature

$$T = K * C_b + T(C_H^{max}) \; , \tag{7}$$

where $K$ is a tunable parameter and $C_b$ is the current best cost. We can also adopt more complicated annealing schedules, such as *adaptive cooling*. The idea is to keep the system close to equilibrium by cooling slower when the specific heat is large. Readers who are interested in different annealing techniques can refer to [9] for more details.

# 4   Experimental Results

To test the ANNEALEDMAP algorithm, we studied its performance on many MAP problems in real Bayesian networks. We compare our results to the algorithms in [19, 20], which are the current state of the art MAP algorithms. We refer to these algorithms as the P-LOC [19] and P-SYS [20] algorithms respectively. We implemented our algorithm in C++ and performed our tests on a 2.4 GHz Pentium IV with 1GB memory Windows XP computer.

## 4.1   Experimental Design

The Bayesian networks that we used include Alarm [5], Barley [14], CPCS [21], Diabetes [2], Hailfinder [1], Munin [3], Pathfinder [11], and Win95pts [12], some of which are constructed for diagnosis. We also tested the algorithms on two very large proprietary diagnostic networks built at the HRL Laboratories (HRL1 and HRL2). The statistics for these networks are summarized in Table 1. Although the network Barley does not have many nodes or arcs, it has a node with 67 states; hence, the P-SYS algorithm failed to give us any results. We also anticipated that Diabetes will be the most difficult for inference algorithms because it has a relatively large ratio between the number of nodes versus the number of arcs.

| Group | Network | #Nodes | #Arcs |
|---|---|---|---|
|   | Alarm | 37 | 46 |
|   | CPCS | 179 | 239 |
| 1 | Hailfinder | 56 | 66 |
|   | Pathfinder | 135 | 195 |
|   | Win95pts | 76 | 112 |
|   | Munin | 1,041 | 1,397 |
| 2 | Barley | 48 | 84 |
|   | Diabetes | 413 | 602 |
| 3 | HRL1 | 1,999 | 3,112 |
|   | HRL2 | 1,528 | 2,492 |

Table 1: Statistics for the Bayesian networks that we are using.

For each network, we randomly generated 20 MAP problems and ran the above three algorithms on them. In each MAP problem, we randomly chose 20 MAP variables among the root nodes or all the root nodes if their number was less than 20. We chose the same number of evidence nodes from the leaf nodes, or all of them if there were fewer than 20 leaf nodes. To set evidence, we sample from the prior of a Bayesian network in its topological order and cast the states of the sample to the evidence nodes.[3]

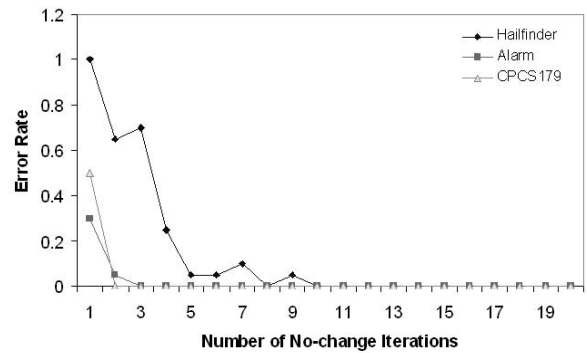

Figure 2: Plot of the error rate against the number of waiting iterations before reheating.

There are several tunable parameters in the ANNEALEDMAP algorithm. We set the initial temperature to be 0.99, the cooling rate of the geometric cooling to be 0.8, and the constant $K$ in Equation 7 to be 0.1 in order that the temperature will not be reset too much above $T(C_H^{max})$. To initialize the Markov chain, we instantiated the MAP nodes sequentially to their most likely states. We did some experiments to select the number of waiting iterations before reheating. Figure 2 is a plot of the error rate against the number of waiting iterations for several Bayesian networks. There is no single number that is optimal for all the networks. We set the number of waiting iterations in all our experiments to be 10 and stopped the algorithm after there had been no improvement for 20 iterations. In other words, we gave the system another chance to reheat before it stopped. For the P-SYS algorithm, we set the time limit to be 50 minutes. For the P-LOC algorithm, we chose the default settings: search=TABOO, initialization=SEQUENTIAL, and maxsteps=25.

## 4.2   Results for the First Group of Networks

In the first experiment, we ran the P-LOC, P-SYS, and ANNEALEDMAP algorithms on all the networks in the first group. The P-SYS algorithm reported that it found all the optimal solutions. Table 2 reports the number of MAP problems that are solved correctly by the P-LOC and ANNEALEDMAP algorithms. They both performed well in these networks. The P-LOC was able to find all the optimal solutions, while the ANNEALEDMAP algorithm missed only one case on the Hailfinder network. The found solution was actually only slightly inferior to the optimal solution: the ratio between their probabilities was 0.997. Search algorithms are able to find optimal solutions because they are deterministic algorithms: they will always choose a better solution if they can find one. However, the ANNEALEDMAP algorithm only selects a solution with some probability. If a solution is only slightly better than another one, the ANNEALEDMAP algorithm will choose the better one with a higher probability, but it is not guaranteed. Our initial experiments show that combining greedy search into ANNEALEDMAP leads to further improvement, which we omit due to page limit.

---

[3]We would like to thank Greg Cooper for this idea.

|            | P-Loc | A-MAP |
|------------|-------|-------|
| Alarm      | 20    | 20    |
| CPCS       | 20    | 20    |
| Hailfinder | 20    | 19    |
| Pathfinder | 20    | 20    |
| Win95pts   | 20    | 20    |
| Munin      | 20    | 20    |

Table 2: The number of cases that are solved correctly out of 20 random cases for the first group of networks.

Besides precision of the results, we also care about the efficiency of the algorithms. Table 3 reports the average running time of the three algorithms on the first group of networks. The median running times are similar. The ANNEALEDMAP, P-LOC and P-SYS algorithms showed similar efficiency on all the networks except the Munin network. The Munin network has many more nodes and arcs than other networks, hence its state space is much larger. The ANNEALEDMAP algorithm drastically reduces the search space by guided random walk, hence it was much faster on the Munin network.

|            | P-Sys  | P-Loc   | A-MAP  |
|------------|--------|---------|--------|
| Alarm      | 0.058  | 0.055   | 0.068  |
| CPCS       | 0.107  | 1.132   | 0.169  |
| Hailfinder | 4.007  | 0.145   | 0.397  |
| Pathfinder | 0.150  | 0.125   | 0.080  |
| Win95pts   | 0.089  | 0.074   | 0.183  |
| Munin      | 50.636 | 193.226 | 21.326 |

Table 3: Average running time in seconds of the P-Sys, P-Loc, and ANNEALEDMAP algorithms on the first group of networks.

## 4.3   Results for the Second and Third Group

The second and third group of networks consist of several large and complex Bayesian networks. The P-SYS did not produce results in a reasonable time for the second group of networks. Therefore, we compared the results of the ANNEALEDMAP algorithm against those of the P-LOC algorithm. Table 4 lists the number of cases that were solved differently between them and the number of cases that the ANNEALEDMAP algorithm produced better results in terms of probability. The two algorithms roughly agreed with each other on the Diabetes network, while differed a lot on the Barley network.

|          | #Different | #A-MAP leads |
|----------|------------|--------------|
| Barley   | 16         | 5            |
| Diabetes | 4          | 2            |

Table 4: The number of cases that are solved differently from P-LOC algorithm by ANNEALEDMAP.

To take a closer look, we report the average and median ratios between the probabilities of the results of the ANNEALEDMAP algorithm and the P-LOC algorithm for the cases that they differed in Table 5. Overall, the P-LOC algorithm performed slightly better on the Barley and Diabetes networks. However, the quality of the MAP solutions were very close.

|          | Ratio | Median |
|----------|-------|--------|
| Barley   | 0.96  | 0.954  |
| Diabetes | 0.996 | 0.997  |

Table 5: Average ratio between the results of ANNEALEDMAP to P-LOC.

For the third group of Bayesian networks, the ANNEALEDMAP algorithm found all the optimal solutions for the randomly generated 20 cases, while the P-LOC algorithm failed to produce any result. The reason why the P-LOC algorithm fails is that it does not use evidence-based pruning while the P-SYS and ANNEALEDMAP algorithms do.[4]   The HRL1 and HRL2 are extremely large two-layer Bayesian networks. When only 20 MAP and evidence nodes are set, the P-SYS and ANNEALEDMAP algorithms will break the networks into pieces by evidence-based pruning and solve them separately. The consequences are that they can both find the optimal solutions efficiently.

Table 6 reports the average running time of the P-SYS, P-LOC and ANNEALEDMAP algorithms on the second and third groups of networks. The results again show that the ANNEALEDMAP algorithm is more efficient on large networks.

## 4.4   Results for Incremental Evidence Test

Out last experiment focused on the robustness of the three algorithms to the number of nodes in the MAP

---

[4]James Park, personal communication.

|          | P-Sys  | P-Loc   | A-MAP   |
|----------|--------|---------|---------|
| Barley   | –      | 87.261  | 45.956  |
| Diabetes | –      | 318.292 | 245.569 |
| HRL1     | 37.034 | –       | 3.503   |
| HRL2     | 28.469 | –       | 8.002   |

Table 6: Average running time in seconds of the P-Sys, P-Loc, and AnnealedMAP algorithms on the second and third groups of Bayesian networks.

set and the evidence set. In this experiment, we generated MAP problems with an increasing number of MAP and evidence nodes and ran the P-Sys, P-Loc, and AnnealedMAP algorithms on these cases. We chose the Munin network for this experiment because only this network has suitable numbers of root nodes and leaf nodes, 183 and 259 respectively, and we were able to run all three algorithms on it. The P-Sys reported that it found all the optimal solutions before it broke down after we set more than 130 MAP and evidence nodes. The P-Loc algorithm found the same solutions before it began to produce results with probability 0 after we set more than 110 MAP and evidence nodes. The AnnealedMAP also found the optimal solutions for the cases whose optimal results are available from the P-Sys algorithm. However, the AnnealedMAP algorithm was able to solve all the cases that we generated, even after we set all the root and leaf nodes to be the MAP and evidence nodes respectively. The running time for all the cases are shown in Figure 3. The AnnealedMAP algorithm turned out again to be more efficient than the P-Sys and P-Loc algorithms. It seems that the AnnealedMAP algorithm extends the class of MAP problems that can be solved.

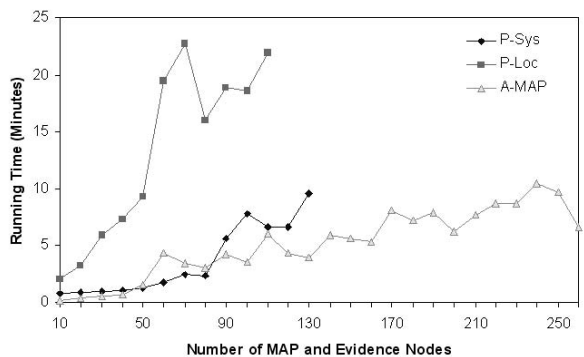

Figure 3: Plot of the running time of the AnnealedMAP, P-Sys, and P-Loc algorithms when increasing the number of evidence nodes on the Munin network.

## 5　Discussion

MAP problems in Bayesian networks are hard because they are not only subject to the complexity of the model (treewidth), but also subject to the complexity introduced by the specific problems (constrained width). The AnnealedMAP algorithm tries to integrate MCMC methods with simulated annealing in order to simulate a non-homogeneous Markov chain that converges to a distribution that concentrates itself on the modes of the target distribution. By taking guided random walks using the Markov chain, the AnnealedMAP algorithm largely reduces the search space of a MAP problem. Our test results show that the AnnealedMAP algorithm is more efficient than the state of the art algorithms, the P-Sys and P-Loc algorithms, on MAP problems in large complex networks. However, the AnnealedMAP algorithm trades off some accuracy for the efficiency. When the problems are not too difficult, the AnnealedMAP algorithm performed slightly worse than the P-Sys and P-Loc algorithms. However, our experiments show that further improvements for AnnealedMAP can be achieved by adjusting the annealing speed, the number of iterations carried out with the same temperature, and the number of iterations before reheating and stopping. Combining greedy search into AnnealedMAP also leads to additional improvement. When the problems become so complex that they are beyond the reach of the P-Sys and P-Loc algorithms, the AnnealedMAP algorithm becomes the only feasible solution. Therefore, the AnnealedMAP algorithm extends the class of MAP problems that can be solved.

## 6　Acknowledgements

This research was supported by the Air Force Office of Scientific Research grant F49620–03–1–0187. We thank Tomek Sowinski and several anonymous reviewers of the UAI04 conference for several insightful comments that led to improvements in the paper, and we thank Adnan Darwiche and Keith Cascio for providing us with an efficient implementation of the P-Sys and P-Loc algorithms within the SamIam software and for their assistance in our tests of SamIam. All experimental data have been obtained using SMILE, a Bayesian inference engine developed at the Decision Systems Laboratory and available at http://www.sis.pitt.edu/~genie.